%% file: 1_Camera_Paper.tex
\documentclass[10pt,twocolumn,letterpaper]{article}
\usepackage[pagenumbers]{wacv}      

\usepackage{graphicx}
\usepackage{amsmath}
\usepackage{amssymb}
\usepackage{booktabs}
\usepackage[pagebackref,breaklinks,colorlinks]{hyperref}
\usepackage[capitalize]{cleveref}
\crefname{section}{Sec.}{Secs.}
\Crefname{section}{Section}{Sections}
\Crefname{table}{Table}{Tables}
\crefname{table}{Tab.}{Tabs.}
\def\wacvPaperID{188} 
\def\confName{WACV}
\def\confYear{2024}
\usepackage{makecell}    
\usepackage{multirow} 
\usepackage[table, dvipsnames]{xcolor}

\input{resources/1_Camera_Tables}

\usepackage{microtype} 
\usepackage{array}
\usepackage{cellspace}
\usepackage{nicematrix}
\usepackage{soul}
\usepackage{arydshln}

\begin{document}
\title{ShadowSense: Unsupervised Domain Adaptation and Feature Fusion for Shadow-Agnostic Tree Crown Detection from RGB-Thermal Drone Imagery}

\newcommand{\authspace}{0.52cm}
\author{
Rudraksh Kapil \hspace{\authspace} Seyed Mojtaba Marvasti-Zadeh \hspace{\authspace} Nadir Erbilgin$^{*}$ \hspace{\authspace} Nilanjan Ray$^{*}$\\
{University of Alberta, Canada}\\
{\tt\small \{rkapil, seyedmoj, erbilgin, nray1\}@ualberta.ca}\\
{\small$^{*}$Equal Contribution}
}
\maketitle
\begin{abstract}
    Accurate detection of individual tree crowns from remote sensing data poses a significant challenge due to the dense nature of forest canopy and the presence of diverse environmental variations, e.g., overlapping canopies, occlusions, and varying lighting conditions. Additionally, the lack of data for training robust models adds another limitation in effectively studying complex forest conditions. This paper presents a novel method for detecting shadowed tree crowns and provides a challenging dataset comprising roughly 50k paired RGB-thermal images to facilitate future research for illumination-invariant detection. The proposed method (ShadowSense) is entirely self-supervised, leveraging domain adversarial training without source domain annotations for feature extraction and foreground feature alignment for feature pyramid networks to adapt domain-invariant representations by focusing on visible foreground regions, respectively. It then fuses complementary information of both modalities to effectively improve upon the predictions of an RGB-trained detector and boost the overall accuracy. Extensive experiments demonstrate the superiority of the proposed method over both the baseline RGB-trained detector and state-of-the-art techniques that rely on unsupervised domain adaptation or early image fusion. Our code and data are available: \href{https://github.com/rudrakshkapil/ShadowSense}{https://github.com/rudrakshkapil/ShadowSense}.
\end{abstract}
\section{Introduction}
\begin{figure}[t]
\centering
    \includegraphics[width=.95\linewidth]{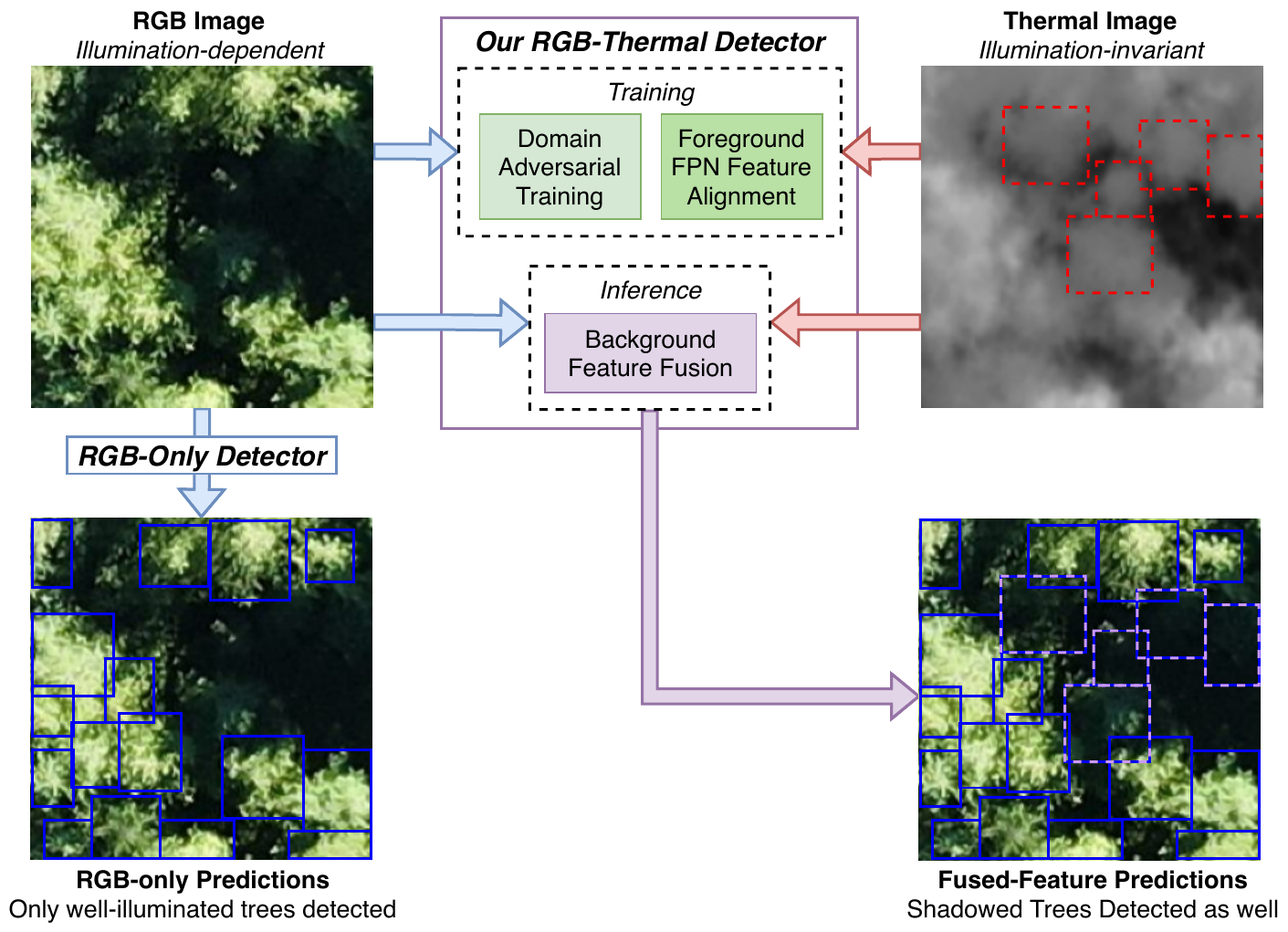}
    \vspace{-0.1cm}
   \caption{\textbf{Overview of Proposed Method}. Undetected trees hidden in shadows are indicated by dotted \textcolor{red}{red} boxes. Best viewed in color.}
\label{fig:challenge}
\vspace{-0.5cm}
\end{figure}
Forest environments are of great importance to ecosystems, economies, and society worldwide. A critical step in forest remote sensing is individual tree crown detection (ITCD), which can assist ecologists, foresters, biologists, and land managers in increasing the scope of their sampling for performing tasks such as forest health monitoring \cite{Ecke2022}, pest infestation detection \cite{Marvasti-Zadeh2022, kapil2022classification}, carbon storage estimation \cite{Fujimoto2019}, and species identification \cite{Beloiu2023, Onishi2021}. 
In recent years, various deep learning-based ITCD methods have been proposed to address the challenges in forest monitoring \cite{Zhao2023}. However, the lack of publicly available, diverse datasets tailored to this specific application has impeded progress in this research domain. Additionally, the ITCD task poses significant application-oriented and environmental challenges. These challenges encompass effectively harnessing the information from multiple sensors and ensuring the robustness of results in the presence of environmental factors. Existing tree crown detectors (e.g., \cite{Weinstein2019}) have primarily been trained on RGB images, which are sensitive to occlusions and illumination variations (e.g., for shorter trees hidden in shadows). Nevertheless, the advantages of incorporating thermal images with complementary information in ITCD have been largely overlooked. 
While a few studies have used RGB-thermal data for urban tree crown detection (e.g., \cite{Moradi2022}), they require extensive manual pixel-wise annotation for supervised training and fail to address forest monitoring challenges, such as shadowed or occluded tree crowns. To bridge these gaps, this paper aims to provide an aligned RGB-thermal forest tree crown dataset and proposes a novel self-supervised approach that leverages both RGB and thermal imagery, improving the accuracy and adaptability of ITCD in various illumination conditions. 

The proposed method (ShadowSense) comprises domain adversarial training (DAT) and foreground (FG) feature alignment to learn domain-invariant representations and match observed tree crowns in both modalities (see~\cref{fig:challenge}). In particular, we train a shadow-agnostic ITCD model consisting of two parallel branches based on the RetinaNet architecture \cite{lin2018focal}. After initializing the branches with RGB-trained weights of the detector, we jointly train the thermal branch and three domain discriminators, minimizing the discrepancy over the feature extractor while maximizing it for domain discriminators. Tree crowns visible in both modalities are adapted by aligning FG feature maps of the feature pyramid network (FPN) using a simple yet effective intensity-based segmentation and morphological operations. During inference, the background (BG) regions of the FPN feature maps from RGB-thermal modalities are fused using a weighted average. The fused maps are then passed to the detector heads, leading to accurate prediction of tree crown bounding boxes. Our proposed method is entirely self-supervised, avoiding the need for labor-intensive manual annotations for model training. Moreover, we provide a challenging large-scale dataset consisting of undistorted and aligned RGB-thermal drone images, serving as a valuable resource to develop robust models and support future research.

The main contributions are summarized as follows. \\
\indent (1) A novel shadow-agnostic tree crown detection method is proposed to exploit complementary information of RGB-thermal images and overcome the limitations of recent RGB-trained models for the ITCD task. This proposed method leverages the registered nature of available data for self-supervision (i.e., eliminating the need for data annotations) and incorporates source domain data post-adaptation. \\
\indent (2) A challenging dataset for shadowed tree crown detection is provided to encompass varying degrees of shadows and illumination conditions of complex forest environments. This RGB-thermal dataset is large-scale and includes annotated images for evaluation and validation, as well as unlabeled images for training, aiming to advance the development of unsupervised/self-supervised methods. \\
\indent (3) Extensive empirical evaluations validate the superior effectiveness of the proposed method over state-of-the-art (SOTA) methods utilizing image-to-image translation, early image fusion, or unsupervised domain adaptation (UDA). 
\section{Related Work \label{sec:Related}}
\textbf{Tree Crown Detection.} 
Deep learning methods have gained significant popularity in ITCD from RGB drone imagery in recent years. These methods primarily rely on well-known object detectors with different architectures \cite{Zhao2023, Hanapi2019} and have found applications in various domains (e.g., \cite{marvastizadeh2023crowncam}). However, these models are often trained on small datasets, resulting in moderate performance and their inability to effectively address challenges like overlapping tree crowns, small crowns, and distractors in various forest environments. Among the existing ITCD methods, DeepForest \cite{Weinstein2019} stands out as the SOTA detector and was trained on a manually annotated dataset comprising over 10k tree crowns from 37 forests across the United States of America. Nevertheless, despite its remarkable performance in well-illuminated conditions, this RGB-only trained detector struggles to accurately detect trees with canopies hidden in shadows.

\begin{figure*}[t]
\centering
\includegraphics[width=.92\linewidth]{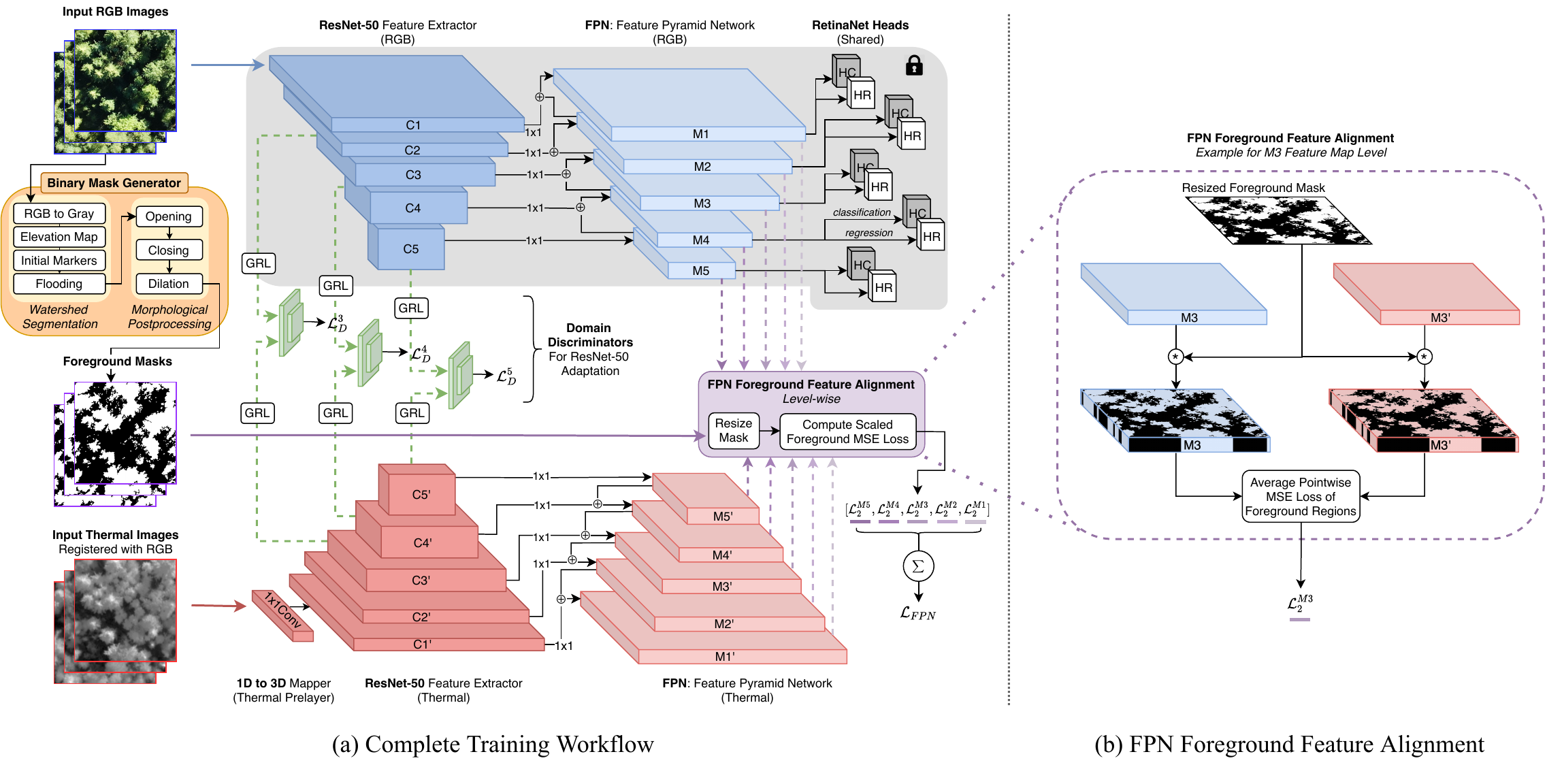}
\vspace{-0.5cm}
   \caption{\textbf{(a) Detailed Workflow of Proposed Training Procedure} consisting of a thermal branch (in \textcolor{red}{red}) and an RGB branch (in \textcolor{blue}{blue}). The weights of both are initialized from \cite{Weinstein2019}, and the latter's are frozen during training. The thermal feature extractor is trained to fool the domain discriminators (in \textcolor{green!50!black}{green}), and vice versa, using gradient reversal layers (GRL) at multiple levels.
   \textbf{(b) Close-up} of FPN feature alignment at the M3 level (in \textcolor{purple!50!blue}{purple}) that encourages foreground feature map regions in the two branches for a given image pair to match.
}
\vspace{-0.2cm}
\label{fig:workflow}
\end{figure*}
\litReviewTable
\textbf{Unsupervised Domain Adaptation (UDA).} 
The goal of UDA is to transfer knowledge from a source domain (e.g., RGB) to a target domain (e.g., thermal) without relying on annotations specific to the target domain \cite{Oza2023}. In general, UDA involves adapting models either within the same modality (e.g., clear vs. foggy RGB) or across different modalities, such as RGB-thermal \cite{Mattolin2022, Vidit2021, Zhao2020, Saito2019, Gan2022, Kim2021, Akkaya2021}. \\
\indent Many UDA approaches incorporate DAT by integrating domain discriminator networks into multiple parts of the model to encourage the learning of domain-invariant representations. These methods often employ global domain classification for the entire image \cite{Pasqualino2021} or local pixel-wise classification, focusing on FG regions \cite{Yang2021} or other areas of interest predicted by attention modules \cite{Kim2021, Vidit2021}. However, the potential application of these methods in the ITCD task is largely unexplored, despite their extensive development and success for generic object detection/segmentation. The drawback of existing UDA methods is their reliance on source domain annotations for training (see \Cref{tab:litReview}), which is not feasible in our problem setting. 
To address this limitation, the proposed method utilizes the registered nature of the available data for self-supervision during adaptation. Moreover, unlike previous approaches, our method retains and incorporates the source domain data after adaptation. 

\textbf{RGB-Thermal Early Fusion.} 
Instead of adapting an RGB-trained model to the thermal data, an alternative approach is to fuse information from both modalities into a more informative image. Most works cope with the lack of ground truth (GT) fusion results by employing unsupervised RGB-thermal fusion methods. These methods can be applied to unregistered or registered images. For instance, UMFusion \cite{Wang2022a} improves upon existing methods for unregistered images by incorporating style transfer and a parallel-branch fusion module. For registered images, Wang \etal \cite{Wang2023} propose an attention-based method for integrating thermal target perception and RGB detail characterization. These methods perform fusion at the image level, resulting in a new richer image with combined properties from both modalities. \\
\indent Alternatively, fusion can be conducted at the intermediate feature level. Moradi \etal \cite{Moradi2022} propose a U-Net-based fusion model for tree crown segmentation, which requires GT segmentation maps for training. Supervised intermediate feature fusion methods have also been extensively studied in tasks like classification \cite{Li2019}, segmentation \cite{Liang2023, Sun2019}, and salient object detection \cite{Wang2022, Song2022, Zhou2022, Liao2022, Tu2021, Guo2021, Zhang2021} (see \Cref{tab:litReview}). In contrast, our method performs feature fusion during inference, rather than early fusion at the image level, in a self-supervised manner specifically designed for ITCD.

\textbf{Image-to-Image Translation.} 
Aside from fusion approaches, an alternative research direction involves colorizing a thermal image to resemble its RGB counterpart using encoder-decoder networks \cite{Luo2022, Chen2021, Huang2018}, and leveraging it for downstream tasks. Another approach is the use of classical algorithms \cite{Finlayson2009} or SOTA deep learning methods \cite{guo2023shadowformer} to translate RGB images into shadow-free versions. However, these methods typically discard the original RGB images in preference of the translated images, which may suffer from image artifacts and potentially contain less semantic information. Instead, our proposed method effectively fuses intermediate features extracted from both modalities after performing the UDA, thereby preserving the complementary information of RGB and thermal modalities.

\section{Proposed Method and Dataset \label{sec:Proposed}}
In this section, ``visible trees" refers to trees seen in both RGB and thermal images, primarily due to sufficient lighting conditions. In addition, ``shadowed trees" are commonly shorter trees that remain hidden by the shadows of neighboring larger trees in the RGB image but become apparent in the thermal image. 
Due to the limitations of the illumination-dependent RGB modality, RGB-trained detectors are ineffective in identifying a significant number of shadowed trees. This is primarily because signals beyond the visible spectrum are undetectable using RGB alone. Hence, our proposed method first adapts the backbone of the existing baseline detector to the thermal data and then fuses extracted features from both modalities during inference.
In the following, we present our proposed method and then introduce an RGB-thermal dataset that facilitates advancements in challenging shadowed tree crown detection and enables the development of robust models for the ITCD task. 

\subsection{Model Architecture and Training}
Our network includes two parallel branches (i.e., RGB and thermal branches) based on the RetinaNet architecture \cite{lin2018focal} for the detection task (see~\cref{fig:workflow}a). Each branch comprises the backbone network, while the classification and regression heads that produce detection outputs are shared. The backbone includes a ResNet-50 network \cite{He2016} that extracts features at multiple resolutions from input images, and an FPN that combines extracted features from multiple levels. We prepend a 1$\times$1 convolutional layer (referred to as pre-layer) to expand thermal input images to three channels before passing them to the backbone network. We initialize the two branches with weights from a pre-trained RGB tree crown detector \cite{Weinstein2019} and freeze the weights of the RGB branch to maintain its performance in the source domain. This is done because these weights effectively identify tree crowns from RGB images but perform poorly for thermal images, as indicated in Table~\ref{tab:quant_results}. Then, we employ DAT and utilize FG FPN feature alignment to adapt the thermal branch to the target domain distribution. Considering the inherent low texture and contrast of thermal data, our training helps the thermal branch provide accurate predictions for visible and shadowed tree crowns. 

\textbf{Domain Adversarial Training (DAT).} 
We employ DAT to train the feature extractor and thermal pre-layer to learn domain-invariant representations. Inspired by \cite{Pasqualino2021}, we incorporate three domain discriminator networks (shown in green in \cref{fig:workflow}) attached to the 3rd, 4th, and 5th levels of the extractor. These CNN classifiers predict the domain label (i.e., RGB or thermal) for the given feature map during training. Each discriminator is preceded by a gradient reversal layer (GRL) \cite{Ganin2015} that acts as the identity function in the forward pass, i.e., $G({\mathbf{x}})={\mathbf{x}}$, but negates gradients in the backward pass. This layer ensures that gradients flowing through the extractor and discriminators are in opposition. This sets the stage for a two-player game: the feature extractor is trained to produce representations whose original domain is indistinguishable by the discriminators, while the discriminators aim to accurately classify the domain labels based on the feature representations.

\indent We use the single-class focal loss to emphasize challenging images during DAT, as,
\begin{equation}
    \mathcal{L}_{D}^c = -(1-p_t)^\gamma log(p_t),
\end{equation}
where c $\in \{3,4,5\}$ is the level of the feature map, $p_t$ is the predicted domain probability, and $\gamma$ controls the diminishing rate of the modulating factor.
A larger weight is assigned to more difficult instances, thereby increasing the importance of these challenging images in the overall loss calculation.
Then, the game is modeled as a min-max optimization,
\begin{equation}
    \min_{\{\theta_d^3, \theta_d^4, \theta_d^5\}} \max_{\{\theta_r, \theta_p\}} \mathcal{L}_{D}^3 + \mathcal{L}_{D}^4 + \mathcal{L}_{D}^5,
\end{equation}
where $\theta_d^c, c \in \{3,4,5\}$ are the parameters of the three domain discriminators, $\theta_r$ are the parameters of the thermal feature extractor, and $\theta_p$ are the parameters of the thermal pre-layer. Unlike typical UDA works \cite{Pasqualino2021, Vidit2021, Munir2021}, we do not combine adversarial loss with a task-aware detection loss due to the lack of source annotations.

\textbf{FPN Foreground Feature Alignment.}
It is crucial for FPN outputs of the RGB and thermal branches to align for the trees that are visible in both modalities (i.e., FG regions). This alignment acts as a proxy for task-specific loss to guide adaptation during training and is vital for the weighted average fusion process during inference (see \Cref{sec:inference}). 
Thus, we can ensure the effective combination of complementary information from both modalities, leading to improved detection performance for shadowed tree crowns. 
\cref{fig:workflow}b illustrates the alignment process for the third FPN feature map level. 
To do so, we first apply a binary BG/FG mask (described below) to the feature maps from the two branches, and then we compute standard average pixel-wise $L2$ loss between the residual values. Accordingly, five loss values denoted as $L_2^f, f \in \{1,2,3,4,5\}$ are obtained. These losses are then combined in a scaled manner, with higher weightage assigned to the larger feature maps using scaling values $\beta^f, f \in \{1,2,3,4,5\}$, i.e.,
\vspace{-0.05cm}
\begin{equation}
    \mathcal{L}_{FPN} = \sum_{f=1}^{5} \beta^f \mathcal{L}_2^f \hspace{.1cm}.
\vspace{-0.05cm}
\end{equation}
\noindent This alignment is complementary to the UDA process – both have the effect of producing the same feature maps at FG regions regardless of the modality. Therefore, $\mathcal{L}_{FPN}$ is used to update the parameters $\theta_f$ of the thermal FPN as well as the preceding parameters $\theta_r$ and $\theta_p$.

To generate the binary masks used to train the detection model, we employ a simple yet computationally efficient method combining classic watershed segmentation \cite{Szeliski2021_Watershed} and mathematical morphology. This approach avoids the complexity of recent methods that utilize auxiliary neural networks for mask prediction. It is particularly suitable for our task because it leverages the assumption that BG pixels (including shadows) are generally darker than those in the FG. 
According to the binary mask generator in Fig.~\ref{fig:workflow}, we transform each RGB image to grayscale. Then, we mark all pixels with intensity $<\frac{20}{255}$ as 1 (representing the darker BG) and those with intensity $>\frac{100}{255}$ as 2 (i.e., brighter FG). 
We are confident about the FG/BG labels for these pixels, while those with intensities in between (initially unmarked) are determined through Meyer's iterative flooding algorithm \cite{Beucher1993}, as implemented in scikit-image \cite{scikit_image, Soille1990_watershed}. 
In this algorithm, an elevation map is computed using Sobel filtering. This map is then `flooded' starting from the defined FG/BG markers. For this, each marked pixel's neighbors are inserted into a priority queue based on gradient magnitude, with enqueue time serving as a tiebreaker favoring the closer marker.
The pixel with the highest priority is extracted, and if its already-marked neighbors share the same marker, it is assigned to that pixel. All unmarked neighbors that are not yet in the priority queue are enqueued.
This flooding procedure iterates until the queue is empty and all pixels are marked as either FG or BG.
After obtaining the initial binary mask, we apply three morphological operations for further refinement. we use 4-connected 3$\times$3 structuring elements (SEs) to perform (1) \textit{opening} to remove errant FG pixels surrounded by BG (2) \textit{closing} to remove errant BG pixels surrounded by FG, 
and (3) \textit{dilation} to pad FG boundaries and maintain FG performance during inference.
Before using the binary mask, we resize it to match the dimensions at each FPN level.

\begin{figure}[b]
\centering
\includegraphics[width=0.8\linewidth]{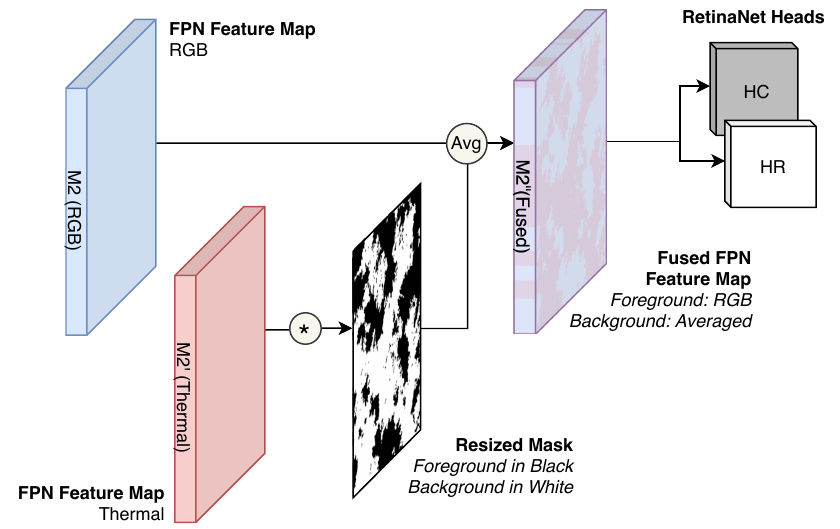}
   \caption{\textbf{Masked Fusion During Inference}, for the M2 level feature maps as an example. Background features (\textcolor{purple!50!blue}{purple}) are obtained by weighted averaging of the RGB (\textcolor{blue}{blue}) and thermal (\textcolor{red}{red}) features. Foreground features are assigned the original RGB values. Best viewed in color.}
\label{fig:inference}
\end{figure}
\subsection{Feature Fusion during Inference} \label{sec:inference}
During the inference phase, we exploit complementary information from the thermal branch to address the limitation of detecting shadowed tree crowns and improve the overall ITCD performance. This information resides in the BG regions of the RGB modality, which are typically prominent in the thermal modality. To achieve this, we follow the binary mask generation process used in the training phase, but now we assign ``1"s to denote BG regions and ``0"s for FG. Subsequently, we fuse the feature maps extracted from the RGB and thermal modalities in a level-wise manner. \cref{fig:inference} illustrates this fusion process for the M2 level of feature maps. 

While the FG pixels (depicted as black regions in~\cref{fig:inference}) from the RGB feature maps are directly utilized, we mask the BG regions of the thermal feature maps to focus solely on the areas that are not visible in the RGB modality. As a result, the fused feature map $F_{Fused}^f$ at level $f$ is obtained through a weighted average of the RGB feature map ($F_{RGB}$) and the thermal feature map ($F_{T}$) for all BG pixels $(x,y)$ as,
\begin{equation}
F_{Fused}^f(x,y) = \frac{F_{RGB}(x,y) + (F_{T}(x,y) \times \lambda_T \times \eta^{f})}{1 + (\lambda_T \times \eta^f)},
\end{equation}
where $\lambda_T$ is the weight assigned to thermal features for all levels and $\eta^f$ denotes the fusion weight scaling specific to that level. $\eta^f$ decreases with $f$ because larger feature maps have a higher spatial resolution, and thus the averaging is less error-prone due to containing more fine-grained information. Once the fused feature map is obtained at each level, it replaces the RGB feature map and is fed into the classification and regression heads to predict bounding boxes.

\datasetsTable

\subsection{Dataset for Shadowed Tree Crown Detection}
In this section, we present an RGB-thermal dataset titled \textit{RT-Trees} for advancing shadowed tree crown detection and developing robust models for ITCD.
We built on an existing dataset \cite{Kapil2023} by conducting additional flights using the same setup. Specifically, we employed a DJI H20T sensor to capture RGB-thermal drone imagery during nine flights over a mixed forested region of central Canada. This data was then combined with available data from the five flights detailed in \cite{Kapil2023}. During data collection, we purposely diversified flight times to encompass a spectrum of challenging illumination conditions. Additionally, ever-changing climatic conditions throughout the year (e.g., temperature and snow cover) introduce an additional layer of diversity and challenges, especially in the more sensitive thermal images. \\
\indent We proceeded with a series of preprocessing steps on the raw drone imagery, including cropping, resizing, co-registration, splitting into training/validation sets based on GPS coordinates, and providing high-quality annotations for evaluation.
This resulted in a substantial collection of approximately 50k registered image pairs across all flights, signifying a considerable expansion compared to existing RGB-thermal datasets (see \Cref{tab:datasets}).
We sampled and annotated 63 non-overlapping images for testing and 10 for validation from a single flight date. Each tree crown only appears once in these sets to ensure the reliability of performance evaluation, and the annotations differentiate between visible and shadowed (i.e., ``difficult") tree crowns.
The remaining bulk of images (49,806) was designated for training. These images display a high degree of overlap ($>75\%$) and span all flights, a deliberate choice aimed at promoting diversity and consequently justifying the discrepancy in data split numbers.
RT-Trees is primarily intended for self-supervised RGB-thermal ITCD, so no training set annotations are provided, but our proposed method demonstrates that the co-registered images can facilitate feature fusion methods. 
A notable characteristic of the RT-Trees dataset is the highly dense spatial distribution of detection targets compared to existing datasets, averaging around 60 tree crowns per image. Moreover, the presence of different tree species results in considerable variability of crown areas and shapes. 
We substantiate the challenges of RT-Trees with descriptive statistics and detail the collection, preprocessing, and annotation procedures in the supplementary material.

\section{Experiments \label{sec:Experiments}}
In this section, we first provide implementation details for the proposed method (ShadowSense). We then compare its performance with the baseline and existing SOTA methods through the quantitative results reported in \Cref{tab:quant_results}. 
We utilize three metrics for evaluation: 
(1) AP50, representing the average precision at 50\% IoU (Intersection over Union) threshold, 
(2) AR100, representing the average recall over several IoUs given 100 detections, and 
(3) Percentage of correctly identified shadowed trees. The third metric focuses only on the difficult boxes, counting a positive if a prediction with an overlap of 85\% with the BG regions was assigned to a difficult box.  
Finally, we present qualitative comparisons to support our experimental findings. 
\subsection{Implementation Details}
We employed the well-known RGB-trained crown detector DeepForest \cite{Weinstein2019} as our baseline method to demonstrate the effectiveness of the proposed method. The RetinaNet networks \cite{lin2018focal} in each RGB/thermal branch were initialized with pre-trained weights from \cite{Weinstein2019}.
To ensure fair comparisons, we configured the RetinaNet hyperparameters similarly to those employed in our baseline. This involved setting the non-maximum suppression threshold to 0.15 and the score threshold to 0.1 (default in \cite{Weinstein2019}). 
During training, we set FPN alignment scales $\beta = [1.0, 1.0, 0.5, 0.05, 0.01]$ and the focal loss parameter $\gamma$ to 2 (recommended in \cite{lin2018focal}).
The domain discriminators consisted of three \textit{Conv-BatchNorm-ReLU-Dropout} layers, an adaptive average pooling layer to reduce feature maps to a single channel, and a linear layer to finally produce a single output representing the confidence 
of belonging to the target domain. Dropout layers with a probability of 0.5 were included for regularization. 
To suppress noisy classification signals during early training stages, we gradually increased the GRL adaptation factor from 0 to 1, as prescribed in \cite{Ganin2015}.
A training batch size of 16 was used in all experiments. The Adam optimizer \cite{adamOptim} was used with an initial learning rate of 0.001 that was exponentially decayed with a gamma factor of 0.9 after each epoch (i.e., a complete pass through the training set). The training was conducted for 10,000 iterations, a sufficient period to observe plateauing in all training losses. The implementations were performed on a single Nvidia GeForce RTX 3090 GPU with 24-GB RAM.
During inference, we performed weighted fusion using a thermal weight of $\lambda_T = 5$, which provided the best results. Similar to $\beta$, the scaling weights $\eta = [1.0, 1.0, 0.5, 0.2, 0.2]$ were applied to weight more towards thermal features in larger feature maps, while also ensuring that all products of $\lambda_T$ and $\eta$ are greater than or equal to one (i.e., always at least equal weighting between thermal and RGB features).
Further validation of selected hyperparameters is provided in the supplementary material.

\resultsTable
\subsection{Baseline Quantitative Comparison}
We evaluated the performance of the baseline model \mbox{\cite{Weinstein2019}} in four scenarios. The first two involved assessing the effectiveness of the off-the-shelf model on RGB images and thermal images, respectively. The performance on RGB images was 49.86\% AP50 and 24.01\% AR100, although only 10.41\% of difficult shadowed trees were successfully identified. When using thermal images, the baseline detector exhibited significantly inferior performance (see \mbox{\Cref{tab:litReview}}). The results demonstrate that the off-the-shelf RGB-trained detector is ill-suited for the thermal domain. In the other two scenarios, we conducted supervised fine-tuning of the detector model on RGB imagery, using supervised focal loss \mbox{\cite{lin2018focal}} for 10 epochs, following \mbox{\cite{Weinstein2019}}. We used a subset of RT-Trees comprising 326 non-overlapping images containing over 22.5k crowns of visible and shadowed trees, which we manually annotated by inspecting the RGB-thermal pairs. 
The performance on RGB images shows a lead of 5.34\% and 5.41\% in terms of AP50 and AP100, respectively, while also resulting in a 9.69\% increase in the detection of shadowed trees. Although the thermal modality is not directly used for training, this configuration requires costly annotation based on both modalities. Also, the performance of this model on thermal images is dramatically poor due to low spatial resolution and lack of fine details in these images. Instead, our ShadowSense can achieve superior performance by leveraging multi-modal data without needing \textit{any} annotations. 

\subsection{State-of-the-art Quantitative Comparison}
We compare the performance of 
ShadowSense with various SOTA methods that utilize image-to-image translation, RGB-thermal early fusion, or UDA. The baseline detector was applied to the generated images in the image translation and fusion methods, and we adopted the proposed weighted-average fusion of BG FPN feature map regions in all UDA experiments (including ours) to ensure fair comparisons. Additionally, an ablation study was conducted to analyze the impact of different components on our method.

\textbf{Image-to-Image Translation.}
We investigated 
three methods: PearlGAN \cite{Luo2022} (SOTA thermal colorization); ShadowFormer \cite{guo2023shadowformer} (SOTA shadow removal); and a classic method that increases the brightness of pixels in HSV color space proportionally to their original brightness, i.e., darker pixels are made brighter. 
Thermal images colorized using PearlGAN performed worse than the baseline by -9.14\% AP50 and -4.34\% AR100. However, slightly more shadowed trees were detected. The decreased performance can be attributed to the introduction of artifacts and an overall loss of semantic information compared to the original RGB images.
The detection performance on images generated by ShadowFormer was the most inadequate.
The classical shadow removal method showed better results than PearlGAN but still performed worse than the baseline. This method jitters the entire image inconsistently with the detector, leading to poor performance.
Translation methods are thus ineffective for removing the shadows of dense canopies in our dataset.

\textbf{RGB-Thermal Early Fusion.}
We evaluated three SOTA methods: UMFusion \cite{Wang2022a}, MFEIF \cite{Liu2022b}, and supervised MetaFusion \cite{Zhao2023}. MetaFusion directly generates a fused RGB image, while UMFusion and MFEIF convert the RGB image to the YCbCr color space, fuse the brightness (Y) channel with the thermal image, and then convert the fused image back to RGB. In all methods, shadowed trees become partially visible in the fused images to varying extents. According to \Cref{tab:quant_results}, MetaFusion performed the best. Although the AP50 and AR100 results were still lower than those of the baseline detector, the fused images revealed 7.59\% more shadowed trees than the baseline. Similar trends were observed for UMFusion and MFEIF. Although RGB-thermal early fusion improved the visibility of BG regions, the detection of tree crowns deteriorated. Overall, the detection performance of all three fusion methods was worse than the baseline.

\textbf{UDA.}
As shown in \Cref{tab:litReview}, existing UDA methods require GT annotations to compute task-specific detection loss during training, which guides the adaptation process. To ensure a fair comparison, we selected three UDA methods compatible with the one-stage object detector RetinaNet, including Attention-based UDA \cite{Vidit2021}, SSTN \cite{Munir2021}, and DA-RetinaNet \cite{Pasqualino2021}. We modified these methods by excluding only the supervised detection loss.
For Attention-based UDA, their proposed attention module, which dynamically selects local feature regions for adaptation, was trained using DAT alongside the thermal branch of our model. In the case of SSTN, only the ResNet and thermal pre-layer were fine-tuned using contrastive loss as described in \cite{Munir2021}. Similarly, for DA-RetinaNet \cite{Pasqualino2021} only global DAT 
was employed to adapt the ResNet and pre-layer of the thermal branch. Among these three methods, DA-RetinaNet demonstrated the best adaptation to the thermal data distribution. However, its performance was still limited as numerous false positive predictions contributed to the overall insufficient performance.
The drawback of these methods lies in the absence of task-aware detection loss during adaptation due to the lack of GT annotations. Consequently, the model cannot learn to extract domain-invariant representations that are meaningful for the detection task. Instead, it primarily learns to deceive the domain discriminators irrespective of the downstream task (i.e., detection). 
Our proposed method overcomes the limitations of the discussed UDA methods by incorporating task-aware FG FPN feature alignment (abbreviated FG FPN FA) to guide the adversarial adaptation process. As a result, our method using DAT and FG FPN FA outperforms the baseline RGB-only detector as well as existing SOTA methods with an AP50 of 54.13\% and AR100 of 25.76\%, with 19.09\% of shadowed trees successfully detected (almost doubling the success rate of the baseline). Particularly, our entirely self-supervised method performs comparably to the supervised fine-tuning method without requiring labor-intensive manual labeling.
The feature fusion process in our method selectively enhances features in the BG regions using thermal-extracted features. Importantly, this fusion does not have an adverse effect on FG performance, which distinguishes our method from existing early fusion approaches. Additionally, our multi-modal fusion leverages available data from both domains for detection, unlike single-domain image-to-image translation methods.
\begin{figure}[b]
\vskip -0.4cm
\centering
\resizebox{0.88\columnwidth}{!}{
    \begin{subfigure}[b]{0.1\textwidth}
         \centering
         \includegraphics[width=1.003\textwidth]{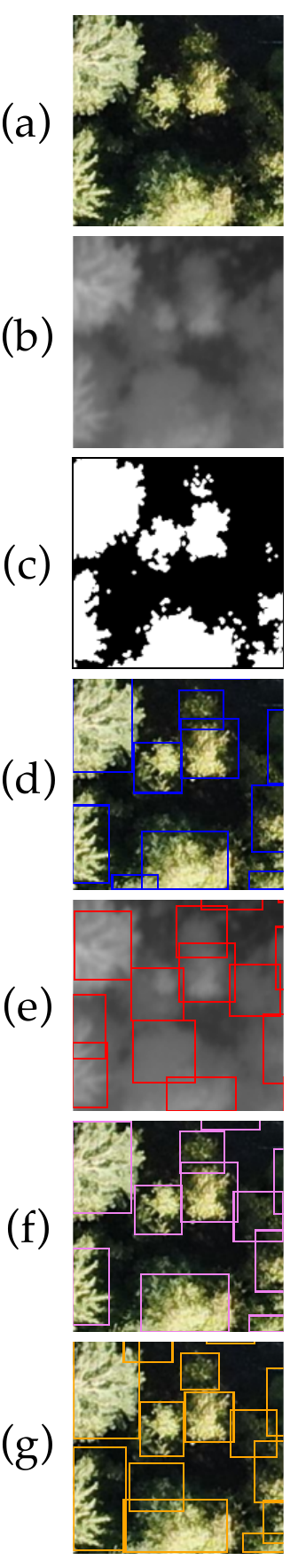}
     \end{subfigure}
     \begin{subfigure}[b]{0.08215\textwidth}
         \centering
         \includegraphics[width=\textwidth]{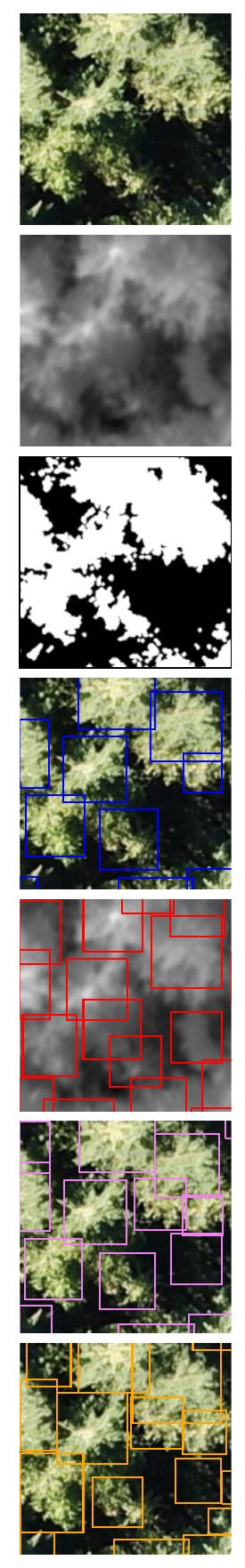}
     \end{subfigure}
     \begin{subfigure}[b]{0.08215\textwidth}
         \centering
         \includegraphics[width=\textwidth]{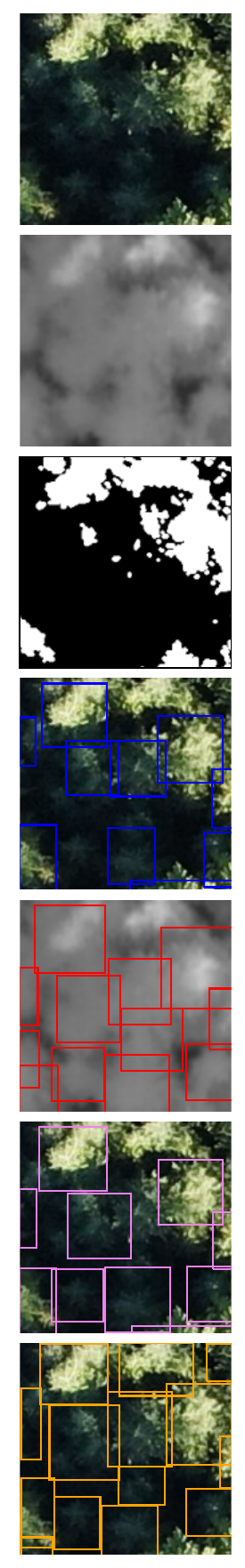}
     \end{subfigure}
     \begin{subfigure}[b]{0.08215\textwidth}
         \centering
         \includegraphics[width=\textwidth]{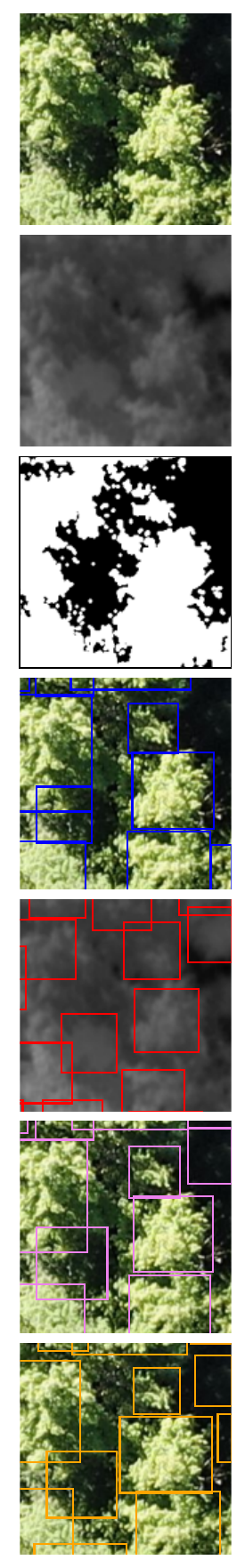}
     \end{subfigure}
     \begin{subfigure}[b]{0.08215\textwidth}
         \centering
         \includegraphics[width=\textwidth]{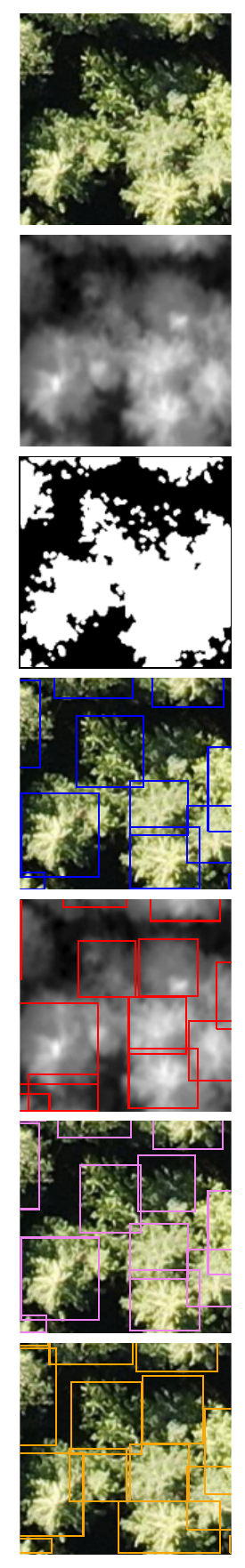}
     \end{subfigure}
     }
\vspace{-0.2cm}    
\caption{\textbf{Detection Results}. Each column shows (a) RGB image, (b) Thermal image, (c) Generated mask; and predictions by (d) Baseline \cite{Weinstein2019}, (e) Our DAT-adapted thermal branch, (f) Proposed ShadowSense, and (e) Ground truth. Best viewed in color and zoom-in.}
\vspace{-0.2cm}
\label{fig:qual_results}
\end{figure}
\subsection{Ablation Study}
A systematic ablation analysis of the proposed method is presented in \Cref{tab:quant_results}. It includes five different configurations: 
(i) FG FPN FA using our classic image masking (CIM) with no DAT applied to the ResNet model,
(ii) ResNet DAT with FPN FA and no masking (aligning all regions of feature maps),
(iii) ResNet DAT with FG FPN FA and a different masking (using baseline detector predictions as FG and the rest as BG),
(iv) Pixel-wise ResNet DAT (discriminators output domain labels for each pixel and consider loss only for FG pixels) with FG FPN FA using CIM, and
(v) Proposed ResNet DAT with FG FPN FA using CIM (ShadowSense). \\
\indent According to the results, the following key inferences can be made:
1) UDA through DAT is crucial (from (i) \& (v)): relying solely on FG FPN FA led to inferior performance compared to the baseline, indicating that the thermal branch did not effectively learn to extract domain-invariant representations without DAT.
2) FG masking for FPN alignment is crucial (from (ii), (iii) \& (v)): Even with DAT, aligning whole feature maps slightly decreased performance compared to the baseline – aligning features of a tree visible only in the thermal image with BG features from the RGB image interferes with training.
3) Our mask generation method outperforms RGB detector-predicted mask generation (from (iii) \& (v)): The proposed masking detects significantly more shadowed trees than this alternate masking, showing superior performance for FPN FA and fusion. 
4) FG masking is unnecessary for DAT (from (iv) \& (v)): Pixel-wise domain classifiers with loss computation restricted to FG regions resulted in slightly worse performance than global DAT, likely due to the usage of less available data (only FG pixels vs. all pixels) in the same number of training iterations.
\subsection{Qualitative Results}
\textbf{Detection Performance.} The performance of the proposed method is compared to the off-the-shelf RGB detector \cite{Weinstein2019} in \cref{fig:qual_results}. Two outputs for our method are shown: one from thermal FPN feature maps (isolated thermal branch) and the other from the fused feature maps. The thermal branch detects shadowed trees in the BG that were missed by the baseline, but there is a decline in the FG performance. In the fused output, the BG detections are accurately propagated while maintaining the baseline performance in the FG. Overall, our method outperforms the baseline by comprehensively improving the detection results. 
Additional qualitative results can be found in the supplementary material.

\begin{figure}[t]
\centering
\includegraphics[width=0.98\linewidth]{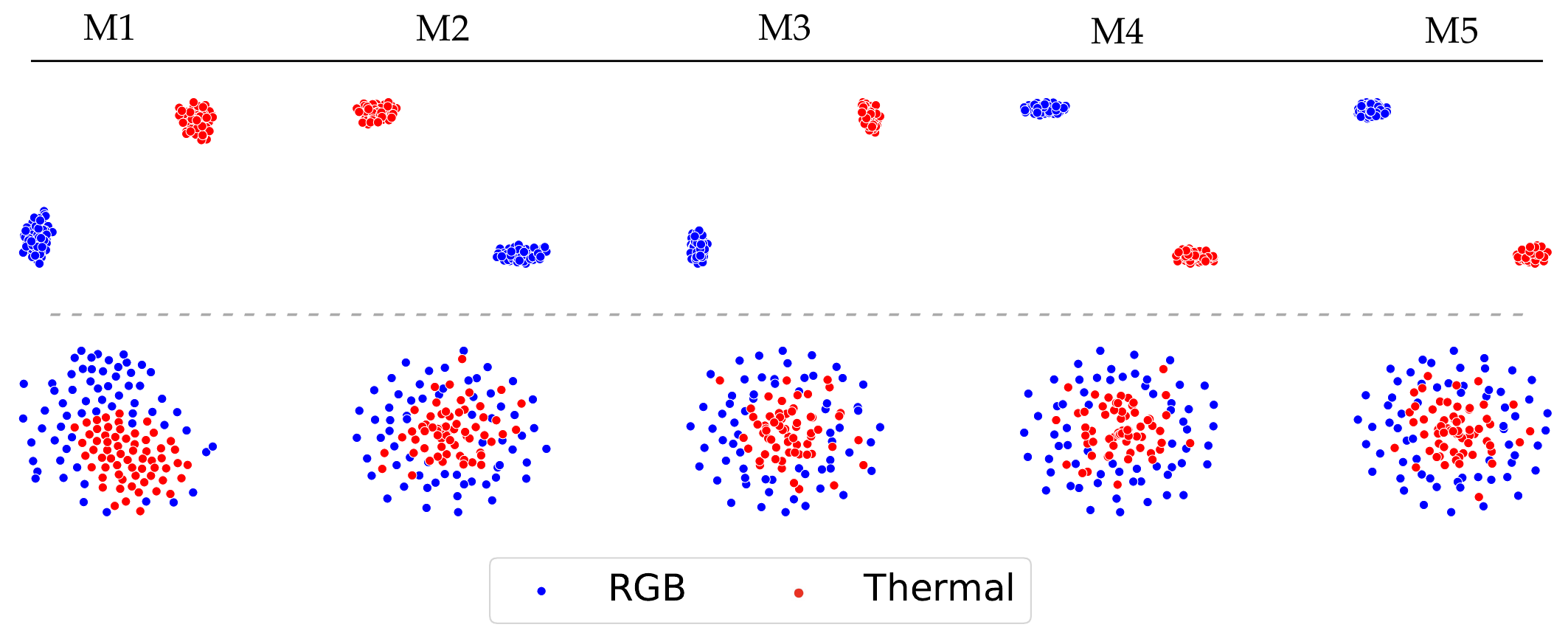}
\vspace{-0.3cm}
\caption{\textbf{t-SNE Visualization} of RGB-thermal FPN features: (top row) before training and (bottom row) after training.}
\vspace{-0.3cm}
\label{fig:feature_vis}
\end{figure}

\textbf{Feature Space Visualization.} 
Visualizing the FPN feature maps for the testing set in \cref{fig:feature_vis} shows the initial disparity between the RGB-thermal modalities before our training procedure. After training, however, they become indistinguishable as domain-invariant feature maps are aligned and can be directly averaged for fusion during inference. 
\section{Conclusions \label{sec:Conclusions}}
We presented a novel shadow-agnostic ITCD method and a challenging paired RGB-thermal dataset to address the limitations of existing RGB-trained detectors. Our method exploits DAT and FG FPN feature alignment to learn domain-invariant representations and match visible tree crowns in RGB and thermal modalities. Unlike existing adaptation methods, our approach does not require source annotations for task-aware supervision during training, but instead relies on the registered nature of image pairs for aligning features of visible FG regions. Our approach effectively detects small trees hidden in the shadow of neighboring taller trees by fusing complementary thermal information. Further, our dataset comprises aligned RGB-thermal drone image pairs that can stimulate future research in challenging ITCD scenarios. Experimental comparisons demonstrate the superiority of our proposed method over the baseline RGB-trained detector and SOTA image fusion- and UDA-based techniques.

\noindent \textbf{Acknowledgments.} We thank fRI Research - MPB Ecology Program for funding; Weyerhauser Company for Cynthia cutblock access; Guillermo Castilla, Mike Gartrell, Jim Weber, and Steven Wagers of NRCan for data acquisition. 

{\small
\bibliographystyle{ieee_fullname}
\bibliography{refs}
}
\end{document}

%% file: resources/1_Camera_Tables.tex
\newcolumntype{M}[1]{>{\centering\arraybackslash}m{#1}}

\makeatletter
\def\igcite#1{%
     \begingroup
         \@fileswfalse
         #1
    \endgroup
}
\makeatother

\newcommand{\litReviewTable}{
\begin{table}[b]
\vspace{-0.15cm}
\centering
\caption{\textbf{Categorization of Related Works} according to training supervision through RGB ground truth (GT) annotations. 
\label{tab:litReview}}
\vspace{-0.2cm}
{\small{
\renewcommand\arraystretch{1.25}
\resizebox{\columnwidth}{!}{
\begin{NiceTabular}{M{0.2\linewidth}|M{0.36\linewidth}|M{0.28\linewidth}}
    \toprule
    \cellcolor{white!20}\textbf{Category} & \cellcolor{red!20}{\textbf{GT Required}} & \cellcolor{green!20}{\textbf{GT Not Required}}\\%
    \midrule
    \midrule

    UDA for Detection
    & 
    \cite{khodabandeh2019robust}
    \cite{Kieu2020}
    \cite{Lyu2022}
    \igcite{\cite{Mattolin2022}} 

    \igcite{\cite{Munir2021}} 
    \igcite{\cite{Pasqualino2021}} 
    \igcite{\cite{Vidit2021}} 
    \cite{Vibashan2022} 
    
    \igcite{\cite{Yang2021}} 
    \igcite{\cite{Zhao2020}}
    \cite{Zheng2020}
    & 
    \textbf{Ours}\\

    \cline{1-3}

    RGB - Thermal Fusion
    & 
    \cite{Gao2022}
    \igcite{\cite{Liu2022a}}
    \cite{Liu2022}
    \igcite{\cite{Moradi2022}}
    \igcite{\cite{Zhao2023a}}

    SOD:
    \igcite{\cite{Guo2021}}
    \igcite{\cite{Liao2022}}
    \igcite{\cite{Song2022}}
    
    \igcite{\cite{Tu2021}}
    \igcite{\cite{Wang2022}} 
    \igcite{\cite{Zhang2021}}
    \igcite{\cite{Zhou2022}}
    & 
    \cite{Han2022}
    \igcite{\cite{Liu2022b}}

    \igcite{\cite{Wang2022a}}
    \igcite{\cite{Wang2023}} 
    \cite{Xu2022}
    
    and \textbf{Ours}\\

    \cline{1-3}
    Translation 
    & 
    \igcite{\cite{Chen2021}}
    \igcite{\cite{guo2023shadowformer}}
    & 
    \igcite{\cite{Huang2018}}
    \igcite{\cite{Luo2022}}\\
    \bottomrule
\end{NiceTabular}}
}}
\end{table}
}

\newcommand{\datasetsTable}{
\begin{table}[t]
\centering
\caption{\textbf{Comparative Overview} of RGB-Thermal Image Datasets.} \label{tab:datasets}
\vspace{-0.2cm}
{\small{
\renewcommand\arraystretch{1.25}
\resizebox{\columnwidth}{!}{
\begin{NiceTabular}{l|c|c|c|c|c}
\toprule
\textbf{Dataset} & \textbf{\# Pairs} & \textbf{Dimensions} & \textbf{Year} & \textbf{GT} & \textbf{Application}\\
\midrule
\midrule
TNO \cite{Toet2022} & 63 & Various & 2014 & $\times$  & Image Fusion\\
MFNet \cite{Ha2017MFNetTR} & 1569 & 640$\times$480 & 2017 & \checkmark  & Semantic Segmentation\\
VIFB \cite{Zhang2020VIFBAV} &  21 &  Various & 2020 & $\times$  & Image Fusion \\
RoadScene \cite{Xu2020} & 221 & 768$\times$576 & 2020 & $\times$  & Image Fusion\\
LLVIP \cite{jia2021llvip} &  15488 & 1080$\times$720 & 2021 & \checkmark  & Pedestrian Detection\\
M$^3$FD \cite{Liu2022a} & 4200 & 1024$\times$768 & 2022 & \checkmark  & Object Detection\\
\textbf{RT-Trees (Ours)} &  49879 &  500$\times$500 & 2023 &  \checkmark (eval.)  & Tree Crown Detection \\
\bottomrule
\end{NiceTabular}}
}}
\vspace{-0.5cm}
\end{table}
}

\newcommand{\resultsTable}{
\begin{table*}[t]
\caption{\textbf{Quantitative Comparison} of the proposed method with baseline and SOTA methods based on AP50 and AP100 metrics.
The best and second-best results are emboldened in \textcolor{red}{red (supervised)} and \textcolor{blue}{blue (self-supervised)}. \label{tab:quant_results}}
\vspace{-0.2cm}
\centering
{\small{
\renewcommand\theadfont{\bfseries}
\renewcommand\theadgape{\Gape[0pt][0pt]}
\renewcommand\arraystretch{1.15}
\resizebox{0.9\linewidth}{!}{
\begin{NiceTabular}{p{3.35cm}p{0.15cm}|l|M{2.5cm}|cc|c}
    \toprule
    \multicolumn{2}{l}{\multirow{2}{*}{\thead{Evaluation}}}
    & \multirow{2}{*}{\thead{Method}}
    & \multirow{2}{*}{\thead{\centering Training Data}}
    & \multicolumn{2}{c}{\thead{All Trees}}
    & \multicolumn{1}{c}{\thead{Shadowed Trees}} \\ \cmidrule{5-7}
    & & & & \thead{\% AP50 $(\uparrow)$} & \thead{\% AR100 $(\uparrow)$} & \thead{\% Identified $(\uparrow)$} \\%
    
    \midrule
    \midrule
    \multirow{4}{*}{\textbf{Baseline \cite{Weinstein2019}}}  
    & & 
    Off-The-Shelf Model (Eval. on RGB Images)
    & \multirow{2}{*}{\shortstack[c]{\parbox{\hsize}{\centering Pre-trained \\on NEON \cite{NEON}}}}
    & 49.86 
    & 24.01 
    & 10.41 \\
    & & 
    Off-The-Shelf Model (Eval. on Thermal Images)
    & 
    & 4.34 & 2.27 & 2.82\\
    \cmidrule{3-7}
    & & 
    Supervised Fine-Tuned Model (Eval. on RGB Images)
    & \multirow{2}{*}{\shortstack[c]{\parbox{\hsize}{\centering + Ann. RGB subset\\of RT-Trees}}}
    & \textbf{\textcolor{red}{55.20}} & \textbf{\textcolor{red}{29.42}} & \textbf{\textcolor{red}{20.10}} \\
    & & 
    Supervised Fine-Tuned Model (Eval. on Thermal Images)
    & 
    &  5.64 &  3.98 & 3.81 \\

    \midrule
    \multirow{3}{*}{\shortstack[l]{\parbox{\hsize}{\textbf{Image Translation}\\Inference using \cite{Weinstein2019} \\on Generated Images}}}
    & & Increased Background Brightness in HSV Space
    & \centering N/A
    & 42.01 & 20.35 & 10.41 \\
    & & ShadowFormer \cite{guo2023shadowformer}: Shadow Removal 
    & \centering ISTD \cite{Wang2018_ISTD}
    & 11.62 & 6.48 & 3.47 \\
    & & PearlGAN \cite{Luo2022}: Thermal Image Colorization 
    & \centering RT-Trees
    & 40.72 & 19.67 & 11.18  \\
    
    \midrule
    \multirow{3}{*}{\shortstack[l]{\parbox{\hsize}{\textbf{Early Image Fusion}\\Inference using \cite{Weinstein2019} \\on Generated Images}}} 
    & & UMFusion \cite{Wang2022a}
    & \centering TNO \cite{Toet2022}
    & 38.00 & 18.56 & 10.20 \\
    & & MFEIF \cite{Liu2022b} 
    & \centering TNO \cite{Toet2022}
    & 39.62 & 18.95 & 15.40  \\
    & & MetaFusion \cite{Zhao2023a} 
    &\centering M$^3$FD \cite{Liu2022a}
    & 43.17 & 21.29 & 18.00 \\

    \midrule
    \multirow{8}{*}{\shortstack[l]{\parbox{\hsize}{\textbf{RGB-Thermal\\ Unsupervised Domain\\ Adaptation (UDA) w/o\\Source Annotations}\\Our Fused Inference \\after Adaptation}}}  
    & & SSTN \cite{Munir2021}: Contrastive Learning 
    & \centering RT-Trees
    & 31.53 & 15.16 & 2.13 \\
    & & Attention-based UDA \cite{Vidit2021} 
    & \centering RT-Trees & 31.97 & 15.34 & 3.84\\
    & & DA-RetinaNet \cite{Pasqualino2021}: ResNet DAT 
    & \centering RT-Trees &  32.88 & 15.72 & 3.65 \\ 
    \cmidrule{3-7}
    & \multirow{5}{*}{\rotatebox{90}{\textit{Ablation Study}}} 
    & $\>$$\-$ (i) \textbf{Ours}: FG FPN FA & \centering RT-Trees & 47.11 & 22.48 & 5.21  \\
    & & $\-$ (ii) \textbf{Ours}: ResNet DAT + FPN FA w/o Masking & \centering RT-Trees & 49.75 & 23.22 & 10.63 \\
    & & (iii) \textbf{Ours}: ResNet DAT + FG FPN FA (Pred. Masks) & \centering RT-Trees & 52.18 & 24.84 & 9.33 \\
    & & (iv) \textbf{Ours}: ResNet FG DAT + FG FPN FA & \centering RT-Trees & 52.24 & 24.38 & 14.32 \\
    & & $\-$ (v) \textbf{Ours} (\textbf{ShadowSense}): ResNet DAT + FG FPN FA & \centering RT-Trees & \textbf{\textcolor{blue}{54.13}} & \textbf{\textcolor{blue}{25.76}} & \textbf{\textcolor{blue}{19.09}} \\
    \bottomrule
\end{NiceTabular}}
}}
\vspace{-0.28cm}
\end{table*}
}

